\documentclass[letterpaper, 10 pt, conference]{ieeeconf}  % Comment this line out if you need a4paper

\IEEEoverridecommandlockouts                              % This command is only needed if 
\usepackage{graphics} % for pdf, bitmapped graphics files
\usepackage{epsfig} % for postscript graphics files
\usepackage{mathptmx} % assumes new font selection scheme installed
\usepackage{times} % assumes new font selection scheme installed
\usepackage{amsmath} % assumes amsmath package installed
\usepackage{amssymb}  % assumes amsmath package installed
\usepackage{cite}
\usepackage{mathrsfs}
\usepackage{color}
\usepackage{booktabs}
\usepackage{makecell}
\usepackage{multirow}
\usepackage{subfigure}

\title{\LARGE \bf
InteractionNet: Joint Planning and Prediction for Autonomous Driving with Transformers
}

\author{Jiawei Fu$^{1}$, Yanqing Shen$^{1}$, Zhiqiang Jian$^{1}$, Shitao Chen$^{1\dag}$, Jingmin Xin$^{1}$, and Nanning Zheng$^{1\dag}$% <-this % stops a space
\thanks{*This work was supported by National Key Research and Development Program of China (SQ2022YFB2500007).}% <-this % stops a space
\thanks{$^{1}$J. Fu, Y. Shen, Z. Jian, S. Chen, J. Xin, and N. Zheng are with the Institute of Artificial Intelligence and Robotics, Xi'an Jiaotong University, Xi'an, Shaanxi 710049, P.R. China;
        {\tt\small fujiawei0724@gmail.com; qing1159364090, flztiii@stu.xjtu.edu.cn; chenshitao@xjtu.edu.cn; jxin, nnzheng@mail.xjtu.edu.cn}}%
\thanks{S. Chen$^\dag$ and N. Zheng$^\dag$ are the corresponding authors.}
}

\begin{document}

\maketitle
\thispagestyle{empty}
\pagestyle{empty}

%%%%%%%%%%%%%%%%%%%%%%%%%%%%%%%%%%%%%%%%%%%%%%%%%%%%%%%%%%%%%%%%%%%%%%%%%%%%%%%%
\begin{abstract}

Planning and prediction are two important modules of autonomous driving and have experienced tremendous advancement recently. Nevertheless, most existing methods regard planning and prediction as independent and ignore the correlation between them, leading to the lack of consideration for interaction and dynamic changes of traffic scenarios. To address this challenge, we propose InteractionNet, which leverages transformer to share global contextual reasoning among all traffic participants to capture interaction and interconnect planning and prediction to achieve joint. Besides, InteractionNet deploys another transformer to help the model pay extra attention to the perceived region containing critical or unseen vehicles. InteractionNet outperforms other baselines in several benchmarks, especially in terms of safety, which benefits from the joint consideration of planning and forecasting. The code will be available at https://github.com/fujiawei0724/InteractionNet. 

\end{abstract}

%%%%%%%%%%%%%%%%%%%%%%%%%%%%%%%%%%%%%%%%%%%%%%%%%%%%%%%%%%%%%%%%%%%%%%%%%%%%%%%%
\section{Introduction}

\begin{figure}[t]
    \setlength{\abovecaptionskip}{0cm}

    \centering
    \includegraphics[width=0.95\columnwidth]{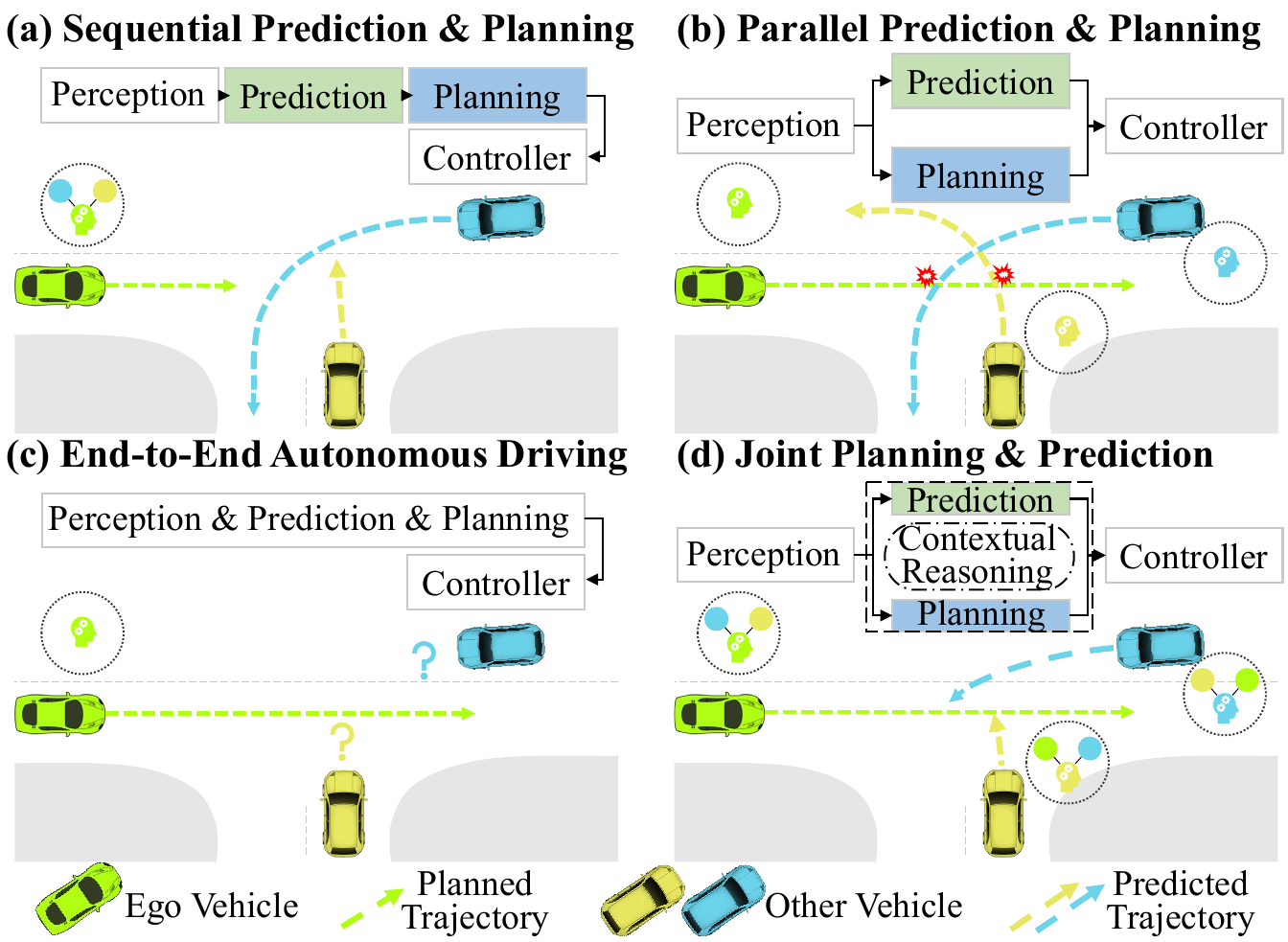}
    \caption{Existing frameworks are divided into three families. Sequential prediction and planning \textbf{(a)} tends to result in overly conservative plans, parallel prediction and planning \textbf{(b)} may generate high collision risk, and end-to-end framework \textbf{(c)} fails to account for future trajectories of other vehicles. We propose joint planning and prediction \textbf{(d)}, which deploys transformer to share global contextual reasoning to handle interaction from all vehicles.}
    \label{intro_figure}
\end{figure}

Autonomous driving has undergone significant progress in recent years, yet its performance still falls short of expert human drivers. One major hurdle is the occurrence of scenarios with multiple traffic participants, which necessitates considering interaction through cooperative planning and prediction. As shown in Fig. \ref{intro_figure}, many researchers \cite{casas2021mp3,sadat2020perceive} employ sequential modules where prediction serves as the upstream module for planning, or they assume parallelism \cite{chen2022learning} between the two modules without considering their interdependence. Additionally, end-to-end methods \cite{chitta2022transfuser, prakash2021multi} are inadequate for accurately accounting for the relationship between planning and prediction, even overlooking it entirely.

In highly interactive driving scenes, expert human drivers tend to incorporate the forecast trajectories of other vehicles into their own planning process. Conversely, when non-ego vehicles produce their future trajectories, they consider the planned trajectories of other vehicles from their own standpoint. All the traffic participants share the global contextual reasoning to achieve interaction and safe dynamic changes in traffic scenarios. For the ego vehicle, answering the contextual reasoning is crucial for addressing adaptation to the dynamic changes of traffic scenarios, and high performance in planning and forecasting. The awareness of global contextual reasoning confers a significant advantage on human drivers in dynamic and complex traffic scenarios.

Motivated by the prior knowledge of expert human drivers, we leverage the attention and information fusion mechanism of transformer \cite{vaswani2017attention} to help the planning and prediction modules integrate global contextual reasoning from each other and coordinate more tightly. Fig. \ref{intro_figure} shows the distinction between existing frameworks and ours. We make the inputs of planning and prediction interconnect with each other through the transformer, which aims to share features between the two modules and reason about the interaction to achieve joint planning and prediction. We term the final model as InteractionNet.

InteractionNet first builds a perception module with semantic segmentation and gets related features for all vehicles through PointPainting\cite{vora2020pointpainting} and PointPillars\cite{lang2019pointpillars} with the assistance of differentiable warping. For the related map-view feature for each vehicle, InteractionNet deploys the local transformer to distinguish the importance of the region in the feature to the future trajectory of the vehicle, which aims to guide the model to give more attention to the area that includes key or unseen vehicles and contains high interactive demand. Then InteractionNet applies the global transformer to share global contextual features among all vehicles and reasons the correlation within planning and prediction and the interaction within all traffic participants. Finally, GRU\cite{cho2014learning} is deployed to generate future trajectories for all vehicles and reach joint planning and prediction.

InteractionNet outperforms the same class methods on several public benchmarks in CARLA simulator \cite{dosovitskiy2017carla}, i.e., Longest6, Town05 short, and Town05 long. We analyze that the provided safety driving and low frequency of infraction penalties of InteractionNet benefit from our joint planning and prediction design. Besides, we do the ablation study to verify the performance of the local and global transformers and get expected and consistent results. 

The contributions of the study are as follows:
\begin{itemize}
    \item We propose InteractionNet, a joint planning and prediction method, which reasons the correlation between the two modules and ranks first place in several benchmarks, especially from the safety dimension.
    \item We apply the global transformer to interconnect and fuse features between planning and prediction to achieve joint planning and prediction.
    \item We use the local transformer to distinguish the importance of the map-view feature from their region and improve the sensitivity to the region with key or unseen vehicles.

\end{itemize}

\section{Related Work}

\subsection{Planning for Autonomous Driving}

The planning module of unmanned vehicles aims to guide the motion of the vehicle and support the driving tasks. Previous researchers \cite{ding2019safe,eiras2021two,pek2020fail,ding2021epsilon} apply hand-designed rules to stipulate the motion planning of vehicles. Ding \emph{et al.} \cite{ding2019safe} and Pek \emph{et al.} \cite{pek2020fail} make the planning system the downstream module of the prediction and navigate the vehicle based on the deterministic future trajectories of other vehicles. While due to the lack of generalization to the unseen scenarios and large consumption to devise rules, recent researchers resort to data-driven methods \cite{chitta2021neat,chen2020learning,wu2022trajectory,zhang2021end,toromanoff2020end}. Casas \emph{et al.} \cite{casas2021mp3} split the mapless navigation as sequential modular perception, prediction, and planning and get expected results in closed-loop simulation environments. Chen \emph{et al.} \cite{chen2022learning} improve the generalization of learning from other vehicles and used a vehicle-independent perception module to get the features for all vehicles and complete parallel planning and prediction. Chitta \emph{et al.} \cite{chitta2022transfuser} and Wu \emph{et al.} \cite{wu2022trajectory} build the end-to-end models and navigate the unmanned vehicles through sensor data directly. Chitta \emph{et al.} \cite{chitta2021neat} build intermediate attention maps from BEV and apply iteration to generate the final waypoints.

Our approach makes a significant contribution by jointly considering planning and prediction. Specifically, we employ a global transformer to seamlessly integrate data across diverse traffic participants and employ reasoning to model their interactions through the interconnected planning and prediction modules.

\subsection{Prediction for Autonomous Driving}

Trajectory prediction pays attention to forecasting the future motion of other traffic participants. It is quite important to self-driving vehicles since its tight relation with the vehicle's planning module. For autonomous driving tasks, the input representation of trajectory prediction includes both privileged history trajectories \cite{gao2020vectornet,zhou2022hivt,sun2022m2i} and sensors data \cite{liang2020pnpnet,luo2018fast,jiang2022perceive}. Sun \emph{et al.} \cite{sun2022m2i} regress the relation among all agents and complete the marginal prediction based on the calculated relation. Zhao \emph{et al.} \cite{zhao2019multi} encode all histories trajectories in the contextual feature and compute all the future trajectories for different agents with the consideration of their interaction. Liang \emph{et al.} \cite{liang2020pnpnet} apply a unified structure to complete the process from sensor data to multiple trajectory prediction. Jiang \emph{et al.} \cite{jiang2022perceive} add an interaction layer between perception and prediction to answer the influence between predicted vectorized maps and detection. 

The current methods for prediction assume independence from the planning module, leading to the neglect of their correlation, and ultimately, a negative impact on driving performance. To address this issue, InteractionNet has been introduced as a collaborative planning and prediction system that utilizes transformers. Additionally, InteractionNet is not reliant on privileged data, as it achieves joint planning and prediction using raw sensor data from the ego vehicle.

\subsection{Transformer in Autonomous Driving}

Transformer has achieved multiple applications in different tasks of autonomous driving: trajectory prediction \cite{9812060,9768029,he2022multi}, detection \cite{9438625,chen2022persformer,erabati2022msf3ddetr}, and planning \cite{li2022lane,reshef2022planning,chitta2021neat}. Huang \emph{et al.} \cite{9812060} modify the multi-head attention to multi-modal attention and design different attention modes, which reach the interpretability and social interactions in trajectory prediction for self-driving vehicles. Zhang \emph{et al.} \cite{Zhang_2022_CVPR} devise a cross-modal transformer to fuse the LiDAR and RGB information for 3D detection in autonomous driving. Liu \emph{et al.} \cite{liu2022augmenting} apply sequential latent transformers with self-supervised learning objectives to distill the latent scene representation, which boosts the reinforcement learning process for the decision-making of unmanned vehicles.  

Unlike previous implementations of transformers in the single task of autonomous driving, we leverage transformers to integrate two tasks, namely planning and prediction, into a joint framework. Our design incorporates the interaction between these tasks by considering the future trajectories of all traffic participants in a mutual manner.

\section{InteractionNet}

\begin{figure*}[t]
    \setlength{\abovecaptionskip}{0cm}

    \centering
    \includegraphics[width=1.8\columnwidth]{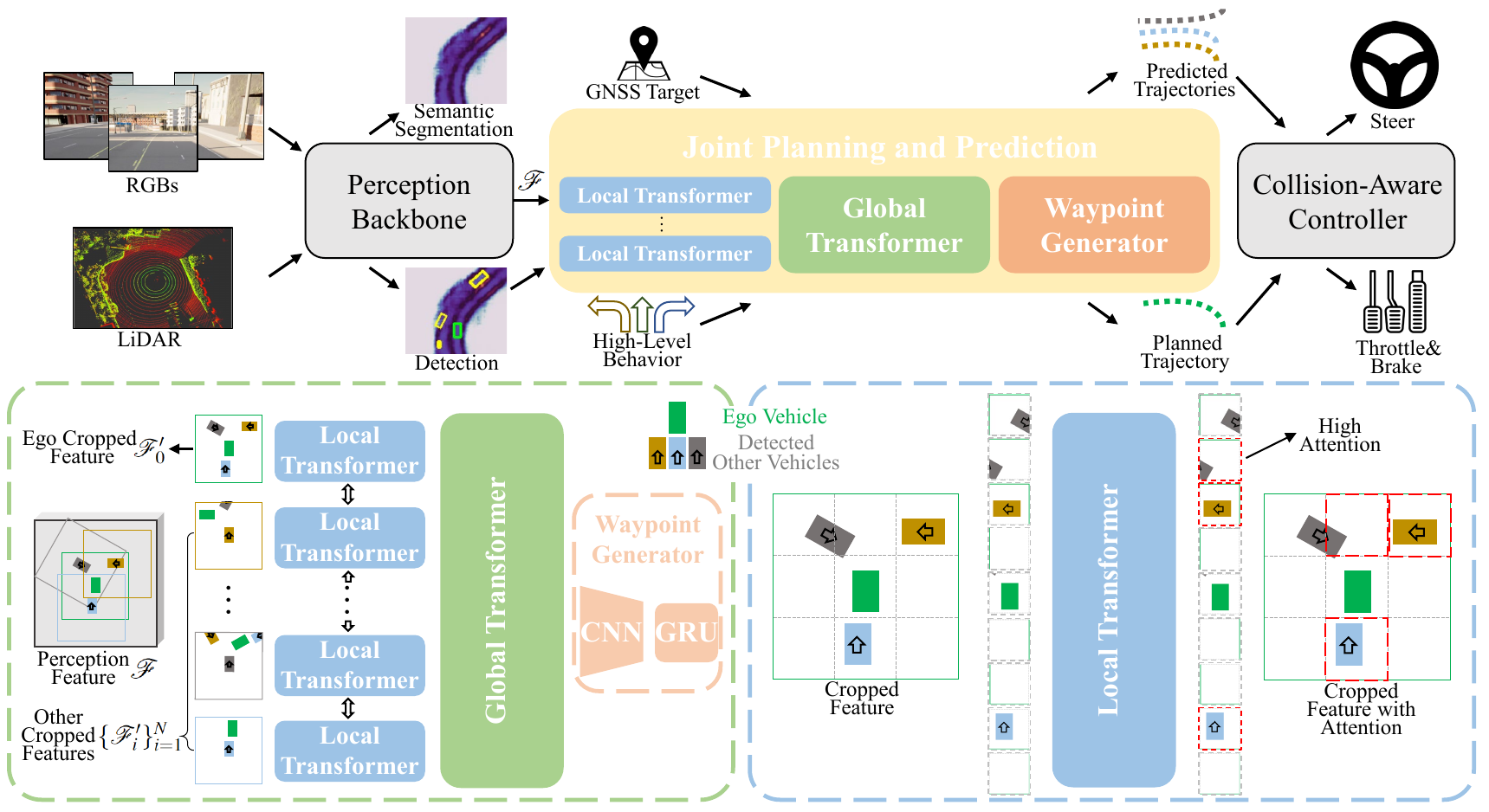}
    \caption{System overview of InteractionNet. It consists of the perception backbone, joint planning and prediction, and collision-aware controller. We deploy a perception backbone with the input of RGBs and LiDAR. The perception backbone provides the detection and map-view perception feature $\mathcal{F}$ to the joint planning and prediction module. The joint planning and prediction aims to provide the future trajectories for both the ego vehicle and other vehicles based on their cropped features $\{\mathcal{F}'_{i}\}_{i=0}^N$ from $\mathcal{F}$. Finally, the collision-aware controller checks the collision risk among all the trajectories and produces the lateral and longitudinal commands to drive the ego vehicle. The detailed architectures of the local and global transformers are shown at the bottom. We feed the local-transformer-processed cropped feature for each vehicle to the global transformer and share the global contextual reasoning among all vehicles, which interconnects the planning and prediction to a joint one. The local transformer aims to guide the model to pay extra attention to distinguish the importance of different regions of the input cropped feature for each vehicle, arising the notice to key or unseen vehicles.}
    \label{system_overview_figure}
\end{figure*}

\subsection{System Architecture}

InteractionNet is designed for mapless navigation and constructs a policy $\pi$ from RGBs, LiDAR, high-level behavior (\emph{go straight, turn left, turn right, following, change left lane, change right lane}), and sparse GNSS target (\emph{2D location}) to the desired throttle, brake, and steer. We apply imitation learning to help build $\pi$ from the CARLA expert, i.e., autopilot. The pipeline of InteractionNet is shown in Fig. \ref{system_overview_figure}. InteractionNet is divided into three closely collaborative parts: perception backbone, joint planning and prediction, and collision-aware controller. The perception backbone is designed to get the map-view perception feature $\mathscr{F}$ from the PointPillars backbone, which is under the supervision of detection and semantic segmentation. The joint planning and prediction transforms $\mathscr{F}$ to the cropped feature $\mathscr{F}'$ for different vehicles from the rotation region based on the ego and detected vehicles' poses. With the superimposition of high-level behavior and coarse GNSS goal, it outputs the trajectories for both planning and prediction. The collision-aware controller checks the safety of the planned trajectory by collision-checking with the predicted trajectories. It also deploys two PID controllers for lateral and longitudinal control and generates the control commands for ego vehicle.  

\subsection{Perception with Semantic Segmentation}

In our mapless navigation context, the perception backbone is responsible to produce the perception feature $\mathcal{F}$, which serves as the input of the joint planning and prediction. $\mathcal{F}$ is also the base of the two tasks. One is detection, i.e., transforming raw sensor data to detection results, e.g., vehicles and pedestrians. The other refers to generating the semantic segmentation results, e.g., road and unavailable area. With two different heads, $\mathcal{F}$ is supervised to provide a comprehensive understanding of the surrounding region, including perceived objects and semantic context.

To this end, similar to LAV \cite{chen2022learning}, we incorporate the CenterPoint \cite{yin2021center} and PointPillars \cite{lang2019pointpillars} as the perception backbone. We also apply PointPainting \cite{vora2020pointpainting} to achieve the fusion of LiDAR and camera inputs. The inputs of the perception backbone include the current LiDAR data $L_{t}$ and three cameras $\mathscr{I}_{t}=\{I_{t}^{1},I_{t}^{2},I_{t}^{3}\}$ with an orientation interval 60°, i.e., one faces the forward, another two orientation in -60° and 60°, respectively. All the cameras have the FOV 120°. The output of our perception backbone is map-view feature representation $\mathscr{F} \in \mathbb{R}^{W \times H \times C}$, where $W,H,C$ mean width, height, and channel, respectively.

In the data preparation stage, the semantic class results required in PointPainting are offered by the trained ERFNet\cite{romera2017erfnet}, which marks semantic information with five classes (\emph{vehicle, pedestrian, road, lane line, unavailable area}). Then we construct the PointNet with two layers of FC with BatchNorm \cite{ioffe2015batch} and ReLU \cite{glorot2011deep} for the PointPillars. The resolution of each pillar is 0.25 $m$ $\times$ 0.25 $m$. We set the valid region of the LiDAR data as abscissa and ordinate in -10 $m$ to +70 $m$ and -40 $m$ to +40 $m$, respectively. We apply CNNs to obtain the final feature $\mathscr{F} \in \mathbb{R}^{384 \times 160 \times 160}$. Note that we sparsify the pillars to improve efficiency in the processing of pillars. For the detection tasks, the CenterPoint detection head provides both vehicles and pedestrians with two centerness maps. For the map-view semantic segmentation, a semantic map is forecast with three choices (\emph{road, lane line, and unavailable area}) based on 3 $\times$ 3 convolution and up-convolution.

\subsection{Joint Planning and Prediction}

The joint planning and prediction is designed to generate the planned trajectory and predicted trajectories simultaneously and consider the correlation between planning and prediction. 

The input of the joint planning and prediction contains perception feature $\mathscr{F}$, the detection, high-level behavior, and sparse GNSS target. We first use the differential warp to rotate $\mathscr{F}$ and locate the regions of interest (RoIs) based on the corresponding pose of each vehicle in the region defined by $\mathscr{F}$, which produces $\mathscr{F}'$ for each vehicle. The outputs are the predicted trajectories $\mathbb{J}=\{\mathscr{J}_{i}\}_{i=1}^{N}$ and planned trajectory $\mathscr{J}_{0}$, where $\mathscr{J}=\{{\bf{p}}_{t}\}_{t=1}^{T}$, $\bf{p}$ denotes the position. 

We design the framework based on a local transformer, a global transformer, and a waypoint generator, see follows.

\noindent \textbf{Local Transformer} is deployed to further process the cropped feature $\mathscr{F}'$ from rotated RoI for each vehicle. The motivation comes from the observation that when human experts aim to calculate the planned trajectory or the predicted trajectories based on the surrounding environment, they must pay special attention to some important regions, i.e., they should use more attention to analyze the regions with key and unseen vehicles or pedestrians, which permeate more interaction and collision risks. We apply the local transformer to distinguish the importance of different regions with the attention mechanism of transformer.  

\begin{figure}
    \setlength{\abovecaptionskip}{0cm}

    \subfigure[The local transformer.] {
    \label{feature_flow_local}     
    \includegraphics[height=0.31\columnwidth]{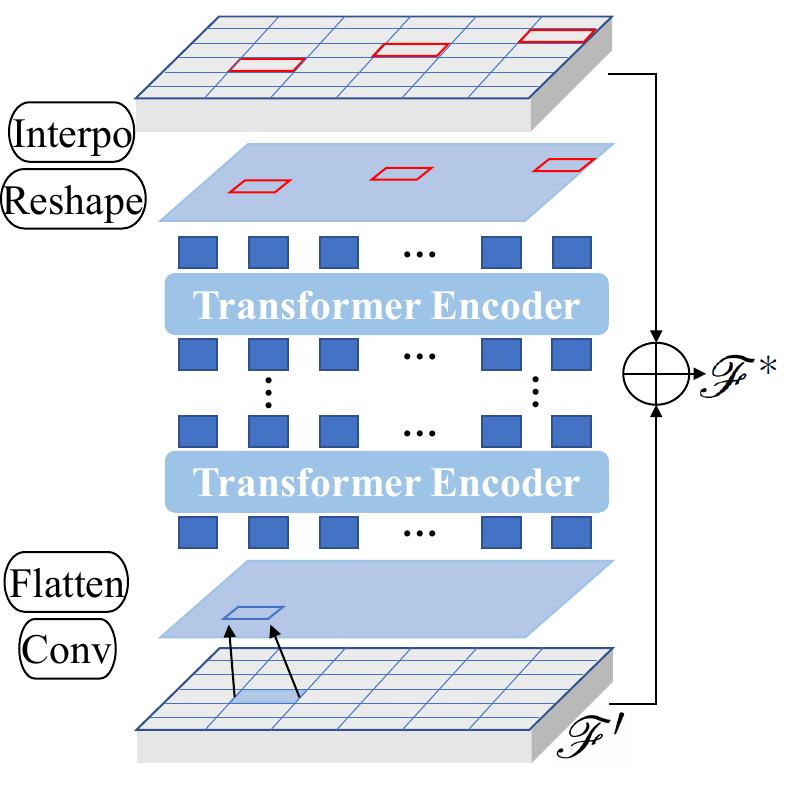}  
    }    
    \hspace{-0.35cm}\subfigure[The global transformer.] { 
    \label{feature_flow_global}     
    \includegraphics[height=0.31\columnwidth]{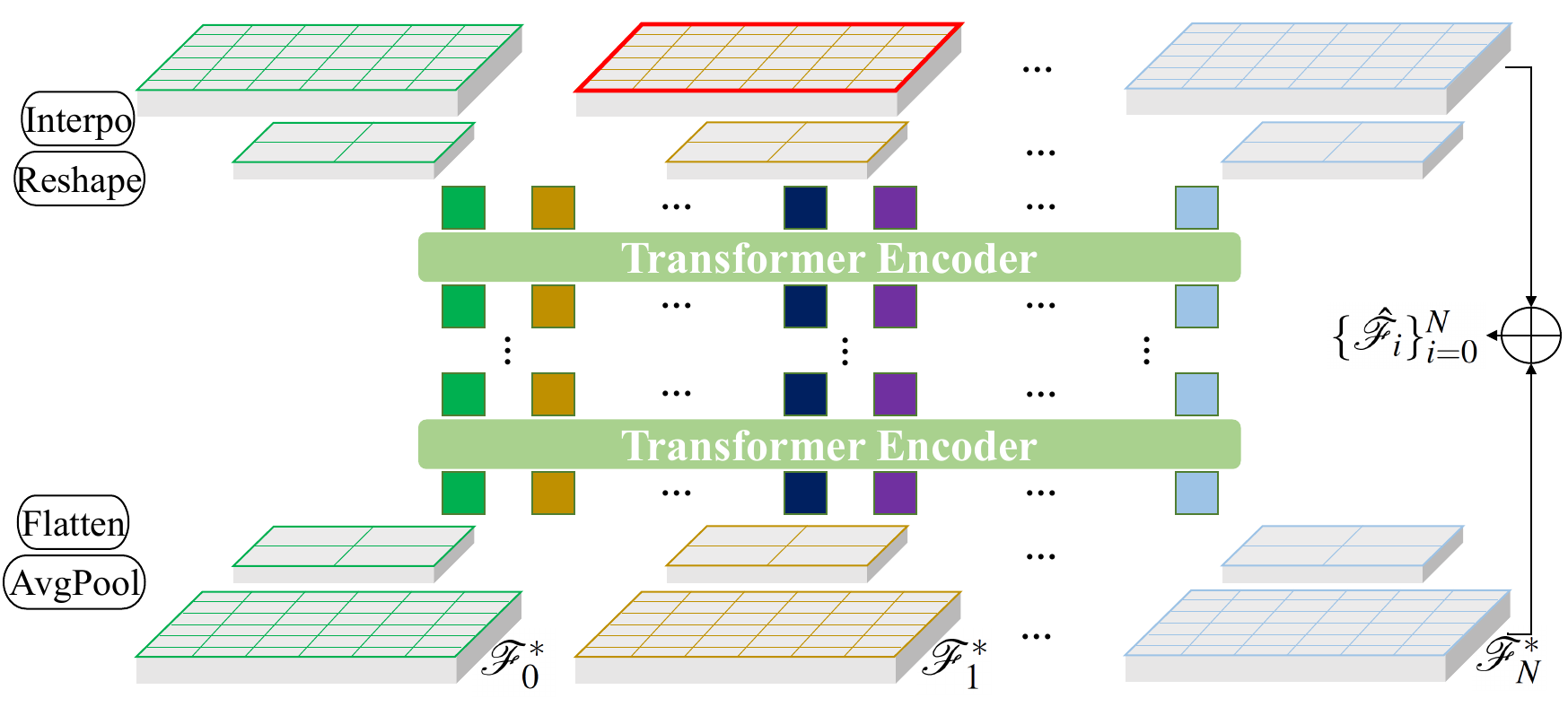}     
    }
    \caption{Feature processing in the local and global transformers. ``$\bigoplus$" is the mathematical addition operation. The local transformer processes the feature for a single vehicle, while the global transformer integrates and shares the features among different vehicles.}
    \label{feature_flow_figure}
\end{figure}

As shown in Fig. \ref{feature_flow_local}, for each $\mathscr{F}'$, we split $\mathscr{F}'$ into 6 $\times$ 6 subfeatures from the width and height dimensions. The split process is realized through a 2D convolution with an equal stride and kernel size. Then we apply the flatten operation to token the split subfeatures to match the input of the transformer. After 6 layers of transformer encoder with 8 attention heads, the processed subfeatures are interpolated to the same shape as the input and superimposed to the original input subfeatures. Finally, we get a processed feature $\mathscr{F}^*$ for each vehicle from the attended subfeatures through mathematical addition operation. We note that the local transformers for both the ego vehicle and detected vehicles share the same parameters. The less the difference between the procedure for planning and prediction, the better the performance to construct a mechanism to share global contextual among all vehicles.

\noindent \textbf{Global Transformer} is designed to integrate global contextual features from all vehicles and interconnect planning and prediction tightly. We select the transformer since it has shown the performance to extract the abstract global feature while considering the interaction among the inputs \cite{zhou2022hivt,renz2022plant}. 

As shown in Fig. \ref{feature_flow_global}, we first use average pooling and flatten each $\mathscr{F}^*$ to match the input of transformers. Then we use 6 transformer encoders layers with 8 attention heads to extract the feature among all the traffic participants. Note that the transformer cannot handle the input sequence of variable length, we set the max input sequence length as 10. For the inference, we select the nearest 9 other vehicles from the detection results and concatenate them with the flattened $\mathscr{F}^*$ of ego vehicle. In the training process, we set a random choice mechanism to select several detected vehicles with a random number maxed to 9. If the number of all the detected surrounding is less than 9, we apply the key padding to mask the placeholder input. Finally, we interpolate and rebuild the cropped feature $\hat{\mathscr{F}}$ for each vehicle through superimpose the output of the global transformer to its input $\mathscr{F}^*$. 

\noindent {\textbf{Waypoint Generator}} focus on generating the corresponding waypoints for all traffic participants. We first apply CNNs and flatten operations to embed $\hat{\mathscr{F}}$ and get a 512-dimensional feature vector for every vehicle. Given the disparity in available information between the ego vehicle and other vehicles, we devise two distinct pipelines to cater to their respective requirements. For ego vehicle, similar to previous methods \cite{chen2022learning, chitta2022transfuser, renz2022plant}, we build several branches according to different high-level behaviors, and every branch corresponds to a GRU. We also deploy a refinement module based on another GRU for the ego vehicle, which considers the GNSS target since its direct relation with the ego planned trajectory. We encode GNSS target and the original planned trajectory into the hidden state of the refinement GRU like Filos \emph{et al.} \cite{filos2020can} and use an MLP to generate the difference between two adjacent waypoints. Finally, we accumulate the calculated difference between every neighboring two waypoints to construct the waypoints. For other vehicles, since we cannot access the high-level behavior and GNSS target, only a CNN and GRU are deployed to generate the waypoints through the accumulation operation.

\subsection{Collision-Aware Controller}

Our controller is able to access future trajectories of all vehicles due to joint planning and prediction. The controller evaluates potential collision risks between the planned trajectory and other predicted trajectories. If there is a high probability of collision, the controller will execute a force stop to ensure safety. This collision-aware approach is crucial for preventing emergent and unsafe situations, see experiments. 

We deploy two PID controllers to generate the lateral and longitudinal commands, i.e., steer, throttle, and brake. In the controller, we find the preview point based on the waypoint that owns the maximum curvature. This way increases the controller's sensitivity to sharp turns. Following LAV \cite{chen2022learning}, we also deploy traffic light and signal detection modules to further enhance driving safety. If the traffic light and signal detection module generate high confidence to stop signals, the ego vehicle will operate force stop. 

\subsection{Loss Function and Learning}

The loss function includes two parts: perception loss and joint planning and prediction loss. The perception loss $\mathscr{L}_{\rm{per}}$ can be written as:
\begin{equation}
    \mathscr{L}_{\rm{per}} = \mathscr{L}_{\rm{det}} + \mathscr{L}_{\rm{seg}}.
\end{equation}
$\mathscr{L}_{\rm{det}}$ is detection loss, which follows the CenterPoint style. $\mathscr{L}_{\rm{seg}}$ is the semantic segmentation loss. $\mathscr{L}_{\rm{seg}}$ is calculated using binary cross-entropy loss for every pixel generated from the simplified semantic segmentation head. While the joint planning and prediction loss $\mathscr{L}_{\rm{jpp}}$ can be written as: 
\begin{equation}
\begin{aligned}
    &\mathscr{L}_{\rm{jpp}} = \mathscr{L}_{\rm{planning}} + \mathscr{L}_{\rm{prediction}},\\
    &\mathscr{L}_{\rm{planning}}=\sum\limits_{t = 1}^T {{{\left\| {{{\bf{p}}_t}|b - {{\bf{p}}_t}^{{\rm{gt}}}} \right\|}_1}},\\
    &\mathscr{L}_{\rm{prediction}} = \sum\limits_{i = 1}^N {\sum\limits_{t = 1}^T {{{\left\| {{\bf{p}}_t^i - {\bf{p}}{{_t^i}^{{\rm{gt}}}}} \right\|}_1}} } .
\end{aligned}
\end{equation}
$\mathscr{L}_{\rm{planning}}$ means the planning loss and $\mathscr{L}_{\rm{prediction}}$ denotes the prediction loss. ${{\bf{p}}_t}|b$ means the generated planned waypoint in time $t$ given high-level behavior $b$, ${{\bf{p}}_t}^{{\rm{gt}}}$ is the ground truth planned waypoint in $t$, $N$ is the number of detected other vehicles, ${\bf{p}}_{t}^{i}$ and ${{\bf{p}}{{_t^i}^{{\rm{gt}}}}}$ means the generated predicted waypoint and ground truth future waypoint for $i_{\rm{th}}$ other vehicles in $t$.  

InteractionNet is a multi-stage autonomous driving system, we train InteractionNet using three steps. In the first step, we only train the perception module with $\mathscr{L}_{\rm{per}}$. In the second step, we freeze the perception module and train the joint planning and prediction model with $\mathscr{L}_{\rm{jpp}}$. In the final step, we train the whole framework with the loss function $\mathcal{L}$:
\begin{equation}
    \mathcal{L}=\mathcal{L}_{\rm{per}} + \lambda \mathscr{L}_{\rm{jpp}},
\end{equation}
where $\lambda$ is the weighting factor to balance the perception loss and joint planning and prediction loss.

\section{Experiments}

We train InteractionNet in all the available 8 towns provided by CARLA. The autopilot in CARLA is the expert to help generate the ground truth data. We use the Adam optimizer \cite{kingma2014adam} with the initial learning rate as 3e-4 and a StepLR scheduler with size as 3, gamma as 0.5 to train InteractionNet. 6 RTX 2080Ti GPUs are operated in the training process with the batch size 30.

\subsection{InteractionNet on Public Benchmarks}
\label{experiment_part_1}

\noindent \textbf{Benchmarks.} We select three public benchmarks to evaluate InteractionNet by comparison with other state-of-the-art methods. \textbf{Longest6} \cite{chitta2022transfuser} contains 36 routes, which are selected from the 6 official public training routes towns. Longest6 chooses the top longest routes given each selected town. The result 36 routes have an average length of 1.5 $km$. Besides, Longest6 tries to keep the density of the participants from insert vehicles and pedestrians at all possible spawn points provided by CARLA. Moreover, Longest6 builds various weather conditions and daylight conditions for different routes. \textbf{Town05 Short} and \textbf{Town05 Long} \cite{prakash2021multi} contain a total of 20 routes. Town05 Short is about 10 short routes about 100 $m$ to 500 $m$ with 3 intersections each. Town05 Long is about 10 long routes about 1000 $m$ to 2000 $m$ with 10 intersections each. Both of them own high densities of dynamic and static agents through the spawn points defined by CARLA. They also have many adversarial scenarios to help distinguish different algorithms.

\begin{table}[t]
    \setlength{\abovecaptionskip}{0cm}

    \setlength\tabcolsep{4.5pt}
    \renewcommand\arraystretch{1.2}
    \centering
    \caption{Comparison results in Longest6 benchmark.}
    \begin{tabular*}{1.0\columnwidth}{c|cccc}
    \toprule 
    \makecell[c]{}&\makecell[c]{\textbf{Method}}&\makecell[c]{\textbf{DS}$\uparrow$}&\makecell[c]{\textbf{RC}$\uparrow$}&\makecell[c]{\textbf{IS}$\uparrow$} \\
    \midrule
    \multirow{4}*{\rotatebox{90}{\makecell[c]{Sensor$\&$\\Mapless}}}&\makecell[c]{WOR\cite{chen2021learning}}&\makecell[c]{20.53$\pm$3.12}&\makecell[c]{48.47$\pm$3.86}&\makecell[c]{0.56$\pm$0.03} \\
    &\makecell[c]{LAV\cite{chen2022learning}}&\makecell[c]{32.74$\pm$1.45}&\makecell[c]{70.36$\pm$3.14}&\makecell[c]{0.51$\pm$0.02} \\
    &\makecell[c]{TransFuser\cite{chitta2022transfuser}}&\makecell[c]{47.30$\pm$5.72}&\makecell[c]{\textbf{93.38}$\pm$1.20}&\makecell[c]{0.50$\pm$0.06} \\
    &\makecell[c]{InteractionNet}&\makecell[c]{\textbf{51.98}$\pm$1.23}&\makecell[c]{87.32$\pm$6.15}&\makecell[c]{\textbf{0.60}$\pm$0.01} \\
    \midrule
    \multirow{5}*{\rotatebox{90}{Privileged}}&\makecell[c]{Rule-Based\cite{renz2022plant}}&\makecell[c]{38.00$\pm$1.64}&\makecell[c]{29.09$\pm$2.12}&\makecell[c]{0.84$\pm$0.00} \\
    &\makecell[c]{PlanT\cite{renz2022plant}}&\makecell[c]{57.66$\pm$5.01}&\makecell[c]{88.20$\pm$0.94}&\makecell[c]{0.65$\pm$0.06} \\
    &\makecell[c]{ROACH\cite{zhang2021end}}&\makecell[c]{55.27$\pm$1.43}&\makecell[c]{88.16$\pm$1.52}&\makecell[c]{0.62$\pm$0.02} \\
    &\makecell[c]{AIM-BEV\cite{hanselmann2022king}}&\makecell[c]{45.06$\pm$1.68}&\makecell[c]{78.31$\pm$1.12}&\makecell[c]{0.55$\pm$0.01} \\
    &\makecell[c]{Expert\cite{chitta2022transfuser}}&\makecell[c]{76.91$\pm$2.23}&\makecell[c]{88.67$\pm$0.56}&\makecell[c]{0.86$\pm$0.03} \\
    \midrule
    \multirow{3}*{\rotatebox{90}{Ablation}}&\makecell[c]{InteractionNet-\uppercase\expandafter{\romannumeral1}}&\makecell[c]{43.08$\pm$2.22}&\makecell[c]{67.25$\pm$3.56}&\makecell[c]{0.58$\pm$0.01} \\
    &\makecell[c]{InteractionNet-\uppercase\expandafter{\romannumeral2}}&\makecell[c]{43.61$\pm$1.73}&\makecell[c]{75.47$\pm$2.16}&\makecell[c]{0.56$\pm$0.04} \\
    &\makecell[c]{InteractionNet-\uppercase\expandafter{\romannumeral3}}&\makecell[c]{40.41$\pm$4.49}&\makecell[c]{72.20$\pm$3.10}&\makecell[c]{0.56$\pm$0.00} \\
    \bottomrule
    \end{tabular*}
    \label{Lonest6_table}
\end{table}

\noindent \textbf{Metric.} We follow the three metrics from the CARLA simulator evaluator. \textbf{Driving Score (DS)} is the main metric, which evaluates the driving performance for an algorithm from the completeness length of the route and both infraction penalties. \textbf{Route Completion (RC)} is the completeness percentage of the agent in each provided route. It evaluates the ability of the models when they face different driving scenarios in different stages of the routes. \textbf{Infraction Score (IS)} records all the unsafe or illegal driving behaviors, e.g., collisions with vehicles and pedestrians, running the stop signals, and red lights. It would accumulate all the behaviors and use predefined weighting factors to calculate the final score.

\begin{table}[t]
    \setlength{\abovecaptionskip}{0cm}

    \setlength\tabcolsep{2pt}
    \renewcommand\arraystretch{1.2}
    \centering
    \caption{Comparison results in Town05 Short and Town05 Long benchmarks.}
    \begin{tabular*}{1.0\columnwidth}{c|ccc|cc}
    \toprule 
    \makecell[c]{}&\makecell[c]{\textbf{Method}}&\multicolumn{2}{c}{\makecell[c]{\textbf{Town05 Short}}}&\multicolumn{2}{c}{\makecell[c]{\textbf{Town05 Long}}}\\
    \midrule
    \makecell[c]{}&\makecell[c]{}&\makecell[c]{\textbf{DS}$\uparrow$}&\makecell[c]{\textbf{RC}$\uparrow$}&\makecell[c]{\textbf{DS}$\uparrow$}&\makecell[c]{\textbf{RC}$\uparrow$} \\
    \midrule
    \multirow{5}*{\rotatebox{90}{Baselines}}&\makecell[c]{CILRS\cite{codevilla2019exploring}}&\makecell[c]{7.47$\pm$2.51}&\makecell[c]{13.40$\pm$1.09}&\makecell[c]{3.68$\pm$2.16}&\makecell[c]{7.19$\pm$2.95} \\
    &\makecell[c]{LBC\cite{chen2020learning}}&\makecell[c]{30.97$\pm$4.17}&\makecell[c]{55.01$\pm$5.14}&\makecell[c]{7.05$\pm$2.13}&\makecell[c]{32.09$\pm$7.40} \\
    &\makecell[c]{AIM\cite{prakash2021multi}}&\makecell[c]{49.00$\pm$6.83}&\makecell[c]{\textbf{81.07}$\pm$5.59}&\makecell[c]{26.50$\pm$4.82}&\makecell[c]{60.66$\pm$7.66} \\
    &\makecell[c]{TransFuser\cite{prakash2021multi}}&\makecell[c]{54.52$\pm$4.29}&\makecell[c]{78.41$\pm$3.75}&\makecell[c]{33.15$\pm$4.04}&\makecell[c]{56.36$\pm$7.14} \\
    &\makecell[c]{InteractionNet}&\makecell[c]{\textbf{66.95}$\pm$2.58}&\makecell[c]{78.42$\pm$4.61}&\makecell[c]{\textbf{40.37}$\pm$0.41}&\makecell[c]{\textbf{72.71}$\pm$1.65} \\
    \midrule
    \multirow{3}*{\rotatebox{90}{Ablation}}&\makecell[c]{InteractionNet-\uppercase\expandafter{\romannumeral1}}&\makecell[c]{53.22$\pm$5.89}&\makecell[c]{74.26$\pm$3.16}&\makecell[c]{34.25$\pm$2.50}&\makecell[c]{65.23$\pm$2.04} \\
    &\makecell[c]{InteractionNet-\uppercase\expandafter{\romannumeral2}}&\makecell[c]{50.84$\pm$3.30}&\makecell[c]{73.99$\pm$4.36}&\makecell[c]{35.78$\pm$2.24}&\makecell[c]{70.29$\pm$4.99} \\
    &\makecell[c]{InteractionNet-\uppercase\expandafter{\romannumeral3}}&\makecell[c]{49.23$\pm$1.07}&\makecell[c]{72.23$\pm$2.36}&\makecell[c]{30.91$\pm$4.31}&\makecell[c]{66.63$\pm$4.09} \\
    \bottomrule

    \end{tabular*}
    \label{Town05_table}
\end{table}

\begin{figure*}[t]
    \setlength{\abovecaptionskip}{0cm}
    \centering
    \includegraphics[width=1.9\columnwidth]{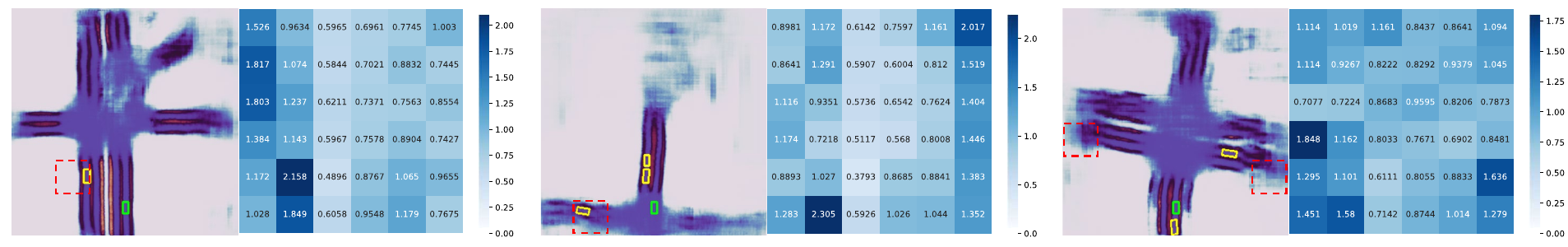}
    \caption{Visualization of the attention scores from the first transformer encoder of the local transformer. The right parts of each subfigure show the accumulated weights for the region among all the queries. The left part corresponds to the predicted map-view semantic segmentation results: \textcolor[RGB]{94,70,167}{road}, \textcolor[RGB]{128,30,80}{lane line}, and \textcolor[RGB]{226,217,226}{unavailable area}, which are calculated from the cropped perceived feature. We also use \textcolor[RGB]{255,0,0}{dashed boxes} to remark the region with high scores in the map-view semantic segmentation results for clarity. The region of the feature that has more attention scores actually is those filled with other vehicles or have more rate to appear unseen vehicles.}
    \label{vis_1_figure}
\end{figure*}

\begin{figure*}[t]
    \setlength{\abovecaptionskip}{0cm}

    \centering
    \includegraphics[width=1.9\columnwidth]{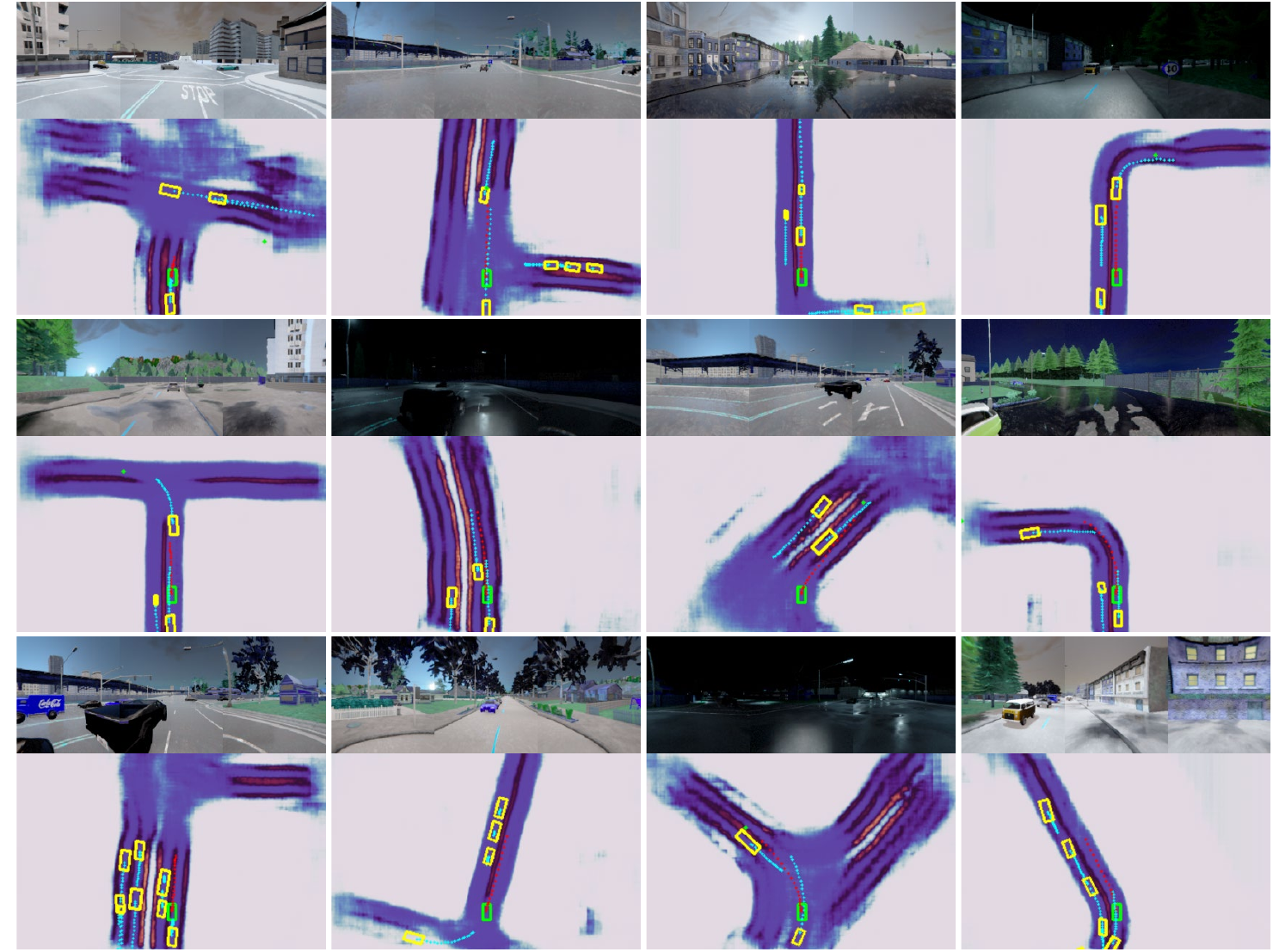}
    \caption{Visualization of the final results. For each subfigure, the top camera pictures are from the deployed left, center, and right cameras, respectively. In the bottom parts, the different semantic segmentation results of the perception backbone are remarked through \textcolor[RGB]{94,70,167}{road}, \textcolor[RGB]{128,30,80}{lane line}, and \textcolor[RGB]{226,217,226}{unavailable area}. The \textcolor[RGB]{0,255,0}{ego vehicle} and \textcolor[RGB]{255,255,0}{other vehicles} and the \textcolor[RGB]{255,0,0}{planned trajectories} and \textcolor[RGB]{41,174,215}{predicted trajectories} are also labeled.} 
    \label{vis_2_figure}
\end{figure*}

\noindent \textbf{Baselines.} For the selected three benchmarks, we choose different baselines since there are missing data on some benchmarks for the specific baselines. \textbf{WOR} \cite{chen2021learning} calculates the action values using dynamic-programming evaluation and distillation for each training process, even if it assumes the world in a rail, the results show its driving performance in the dynamic and reactive world. Note it now is the state-of-the-art method in NoCrash benchmark \cite{codevilla2019exploring} of CARLA. \textbf{LAV} \cite{chen2022learning} builds an intermediate perception module and uses the distillation to teach the agent from a privileged expert. LAV riches the training process of autonomous driving using other vehicles' trajectories and improves its generalization. It builds parallel planning and prediction modules to provide the trajectories of all the vehicles. \textbf{TransFuser} \cite{chitta2022transfuser,prakash2021multi} is an end-to-end autonomous driving system with several auxiliary tasks. It leverages 4 transformers to fuse multi-modal data between LiDAR and RGBs. GRUs are also applied in TransFuser to embed the target from GNSS and generate the final planned trajectory. \textbf{CILRS} \cite{codevilla2019exploring} is a conditional imitation learning method with a single RGB and speed information input. It constructs several independent branches to generate the planned trajectories for different high-level behavior inputs from CARLA. \textbf{LBC} \cite{chen2020learning} gets the ground truth BEV and trains a teacher agent based on the privileged knowledge. Then it uses distillation to train the final agent with the input of 3 input camera RGBs. \textbf{AIM} is a strong baseline provided by Prakash \emph{et al.} \cite{prakash2021multi} to verify the performance of TransFuser. It uses a GRU-based waypoint generation network to replace the corresponding part in CILRS, which encodes the traffic lights and signals to improve the sensitivity of them. We also provide several methods with privileged data input as comparisons. \textbf{Rule-Based} \cite{renz2022plant} method is built with the same input as the autopilot provided by CARLA, the difference locates that Rule-Based cannot access the future trajectories of other vehicles. \textbf{PlanT} \cite{renz2022plant} is an object-level planning system, it uses the state-of-the-art perception module to get the detected objects and uses the attention mechanism of the transformer to process the vector representation of the objects. It accesses the traffic lights precisely from CARLA and the dense future routes using interpolation. \textbf{ROACH} \cite{zhang2021end} applies reinforcement learning to build an agent, which accesses the ground truth BEV with the accurate objects and HD map. \textbf{AIM-BEV} \cite{hanselmann2022king} holds the same input representation as ROACH, but it follows imitation learning framework and uses GRUs with two controllers to produce control commands.

\noindent \textbf{Comparison Results.} Table \ref{Lonest6_table} compares InteractionNet with other baselines in Longest6 benchmark. The performance of the provided baselines is obtained from Chitta's work \cite{chitta2022transfuser}. They repeat each baseline three times and calculate the average value and standard deviation since the consideration of the non-determinism of the CARLA simulator. For fairness, we also test InteractionNet for three times to be parallel with the baselines with a single RTX 2080Ti GPU. We find InteractionNet gets the best DS among all the baselines without privileged information. Even though TransFuser \cite{chitta2022transfuser} gets the best RC, it does not get the better DS since InteractionNet wins in IS, which means InteractionNet avoids more illegal and unsafe situations, e.g., collision with vehicles and pedestrians. We note that this improvement comes from our joint planning and prediction and collision-aware controller. Besides, InteractionNet keeps a large gap between WOR and LAV. InteractionNet even wins in some metrics when compared with the baselines with privileged information. Table \ref{Town05_table} presents the results obtained from the Town05 Short and Town05 Long benchmarks, comparing the performance of InteractionNet with other baselines, as reported in Prakash \emph{et al.} \cite{prakash2021multi}. To ensure a fair comparison, we adopt the same training routes used in the baselines. However, unlike the baselines that only utilize the historical trajectory of the autopilot for training, we use both the historical trajectories of the ego vehicle and other vehicles to train our joint planning and prediction model. Our results demonstrate that for both short and long routes, InteractionNet achieves superior performance (22.7\% and 21.8\%, respectively) compared to other methods in terms of DS. Despite outperforming the other baselines, we still observe some unsafe or struggling scenarios, such as unnecessary collisions, ignoring traffic signals, and getting blocked, particularly in extremely crowded traffic. We attribute these issues to the challenge of processing information from multiple sources simultaneously. To overcome these challenges, we aim to fuse the camera and LiDAR with a more robust and powerful backbone.

\subsection{Ablation Study}

We verify the performance of the local transformer and global transformer, respectively. Therefore, we design three variants of the initial InteractionNet. \textbf{InteractionNet-\uppercase\expandafter{\romannumeral1}} removes the local transformer. \textbf{InteractionNet-\uppercase\expandafter{\romannumeral2}} removes the global transformer. \textbf{InteractionNet-\uppercase\expandafter{\romannumeral3}} removes both the local and global transformers. We also test the three variants in with the same benchmarks in Section \ref{experiment_part_1}. Table \ref{Lonest6_table} and Table \ref{Town05_table} shows the comparison results. InteractionNet-\uppercase\expandafter{\romannumeral1} has some degradation since it loses the mechanism to enhance the more related feature. InteractionNet-\uppercase\expandafter{\romannumeral2} cannot ensure the IS and suffers more collision risks since it does not have the global transformer to interconnect the planning and prediction. InteractionNet-\uppercase\expandafter{\romannumeral3} has the lowest DS and helps verify the performance of the deployed transformers. The full model outperforms all the variants and shows the effectiveness of our modifications. 

\subsection{Qualitative Results}

\noindent \textbf{Visualization of the attention scores.} We show the attention scores of the local transformer in three scenarios in Fig. \ref{vis_1_figure}. We use the cropped perceived feature for the ego vehicle for example. We divide the whole feature into 6 $\times$ 6 tokens, each denoting an area of the perceived region. Since every transformer encoder generates the weights matrix for every tokens pair, which is too large and complicated, the calculated weights in Fig. \ref{vis_1_figure} are the accumulation results from all query features in the first transformer encoder. The results show that the local transformer will give the region more attention scores, which owns the influenced obstacles or has a high probability to be occupied by unseen vehicles. 

\noindent \textbf{Visualization of the joint planning and prediction.} We visualize the final results of InteractionNet in Fig. \ref{vis_2_figure}. The test scenarios include both day and night, sunny and rainy. The visualization illustrates that InteractionNet could complete the joint planning and prediction tasks in different scenarios and also with different traffic participant densities. 

\section{Conclusion}

In this work, we propose a joint planning and prediction framework based on transformers, termed InterctionNet. we employ a local transformer to prioritize the salient areas of the map-view cropped features derived from perception. Moreover, we integrate a global transformer to connect the planning and prediction modules, facilitating a collaborative architecture. Through joint planning and prediction, InteractionNet achieves superior performance in multiple public benchmarks, particularly in reducing collision risks and infraction penalties.


%%%%%%%%%%%%%%%%%%%%%%%%%%%%%%%%%%%%%%%%%%%%%%%%%%%%%%%%%%%%%%%%%%%%%%%%%%%%%%%%

%%%%%%%%%%%%%%%%%%%%%%%%%%%%%%%%%%%%%%%%%%%%%%%%%%%%%%%%%%%%%%%%%%%%%%%%%%%%%%%%

%%%%%%%%%%%%%%%%%%%%%%%%%%%%%%%%%%%%%%%%%%%%%%%%%%%%%%%%%%

\bibliographystyle{IEEEtran}

\bibliography{IEEEexample}

\end{document}